\documentclass[runningheads]{llncs}
\usepackage[T1]{fontenc}
\usepackage{graphicx}

\usepackage{booktabs} 
\usepackage{amsmath,url,graphicx,amssymb}
\usepackage{amsfonts}
\usepackage{multirow}
\usepackage{algorithm}
\usepackage{algpseudocode}
\usepackage{pifont}
\usepackage{color}
\usepackage{xspace}
\usepackage{enumitem}
\usepackage{subcaption}
\usepackage{cite}
\usepackage{colortbl}
\usepackage{xcolor}
\usepackage{bm}
\usepackage{caption} 
\usepackage{todonotes}
\usepackage[
  separate-uncertainty = true,
  multi-part-units = repeat
]{siunitx}

\usepackage{hyperref}
\usepackage{color}
\definecolor{b2}{RGB}{51,153,255}
\definecolor{p2}{RGB}{121,64,255}
\definecolor{kb}{RGB}{185, 60, 248}

\newcommand{\ours}{{$\mathsf{EXAMINE}${}}}
\newcommand{\eg}{{\em e.g.,~}}           
\newcommand{\ie}{{\em i.e.,~}}

\newcommand{\bA}{\boldsymbol{A}}
\newcommand{\bB}{\boldsymbol{B}}

\newcommand{\cL}{\mathcal{L}}
\newcommand{\E}{\mathbb{E}}

\newcommand{\bR}{\mathbb{R}}

\begin{document}
\title{Efficient Medical Image Assessment via Self-supervised Learning}
\titlerunning{Efficient Medical Image Assessment via Self-supervised Learning}

\author{Chun-Yin Huang\inst{1} \and
Qi Lei\inst{2} \and
Xiaoxiao Li\inst{1}}

\institute{University of British Columbia \and Princeton University\\
\email{\{chunyinh, xiaoxiao.li\}@ece.ubc.ca, qilei@princeton.edu}}

%
%
\maketitle              

\begin{abstract}
High-performance deep learning methods typically rely on large annotated training datasets, which are difficult to obtain in many clinical applications due to the high cost of medical image labeling. Existing data assessment methods commonly require knowing the labels in advance, which are not feasible to achieve our goal of \textit{`knowing which data to label.'} To this end, we formulate and propose a novel and efficient data assessment strategy, \textbf{EX}ponenti\textbf{A}l \textbf{M}arginal s\textbf{IN}gular valu\textbf{E} ($\mathsf{EXAMINE}$) score, to rank the quality of unlabeled medical image data based on their useful latent representations extracted via Self-supervised Learning (SSL) networks. Motivated by theoretical implication of SSL embedding space, we leverage a Masked Autoencoder~\cite{he2021masked} for feature extraction. Furthermore, we evaluate data quality based on the marginal change of the largest singular value after excluding the data point in the dataset. We conduct extensive experiments on a pathology dataset. Our results indicate the effectiveness and efficiency of our proposed methods for selecting the most valuable data to label.

\end{abstract}
%
%

\section{Introduction}

Artificial intelligence (AI) such as deep learning has became a powerful tool for medical image analysis. Its success relies on the  availability of abundant high quality dataset. However, medical images collected from different sources vary in their quality due to the various imaging devices, protocols and techniques. When trained with low-quality data, AI models can be compromised. Furthermore, labeling medical images for AI training requires domain experts and is usually costly and time consuming. Therefore, it is demanding to have an automated framework to effectively assess and screen data quality before data labeling and model training.


 There are numerous definitions of data quality. Data is generally considered to be of high quality if ``fit for [its] intended uses in operations, decision making and planning.''~\cite{redman2008data,fadahunsi2019protocol,fadahunsi2021information}. In the context of training an AI predictive model, good data are the fuel of AI. Namely, data with better quality can help obtain higher prediction accuracy. However, how to quantitatively assess data's quality for AI tasks is under-explored. Previous works \cite{jia2019towards, pmlr-v97-ghorbani19c}  mainly propose to estimate data values in the context of supervised machine learning, which requires knowledge of labels and repeated training of  a target utility. Such setting lacks practical value as data labels are typically not available at the data preparation stage for data privacy, labeling cost, and computational efficiency concerns.  
 Differently, we aim to develop a cost-effective scheme for data assessment in the context of unsupervised learning to tackle the limitations of the existing methods, in which no labeling is required during assessment.

A trending and powerful unsupervised representation learning strategy is self-supervised learning (SSL). SSL solves auxiliary pretext tasks without requiring labeled data to learn useful semantic representations. These pretext tasks are created solely using the input features, such as predicting a missing image patch~\cite{he2021masked},
recovering the color channels of an image from context~\cite{zhang2016colorful}, predicting missing words in texts~\cite{devlin2018bert}, forcing the similarity of the different views of images~\cite{chen2020simple, grill2020bootstrap}, etc. 
Motivated by the recent discovery that SSL could embed data into linearly separable representations under proper data assumptions~\cite{lee2021predicting,tosh2021contrastive}, we show that `good' and `bad' data can be distinguished by examining the change of the data representation matrices' singular value by removing a certain data point.

In this work, we tackle a practically demanding yet challenging problem --- medical image assessment (also referred as data assessment in this work). To this end, we develop a novel and efficient pipeline for medical image assessment \textit{without knowing data labels}. As shown in Fig~\ref{fig:pipeline}, we propose a new metric, \textbf{EX}ponenti\textbf{A}l \textbf{M}arginal s\textbf{IN}ular valu\textbf{E} (\ours) score to evaluate the value (or referred as quality) of individual data by first using SSL to extract the features, and then calculate the value of the data using Singular Value Decomposition (SVD). \ours{} scores are useful in indicating the essential data to be annotated, which can not only abundantly reduce the effort in manual labeling but also mitigate the negative effect of mislabeled data, and further improve the target model. Our chief contributions are summarized as follows: \\
    \noindent - We are the first to show the feasibility of using an unsupervised learning framework to assess medical data by utilizing SSL and SVD, which is a more cost-efficient and practical method to evaluate data compared to previous work.\\
    \noindent - \ours{} can assess data \textit{without} knowing the label, which reduces annotation efforts and the chance of mislabeling.\\
    \noindent - We conduct experiments on the simulated medical dataset to demonstrate the feasibility of using \ours{} scores to distinguish data with different qualities and show comparable performance to previous supervised learning based works.

\section{Preliminaries}

\subsection{Supervised-learning-based Data Assessment}
\label{sec:shap}
The goal of data assessment is using a valuation function to map an input data to a single value that indicate its quality.  Supervised-learning-based data assessments assume knowing a labeled dataset $\mathcal{N}^{l}=\{(x_i,y_i)|i\in[N], x_i \in \mathcal{X}, y_i \in \mathcal{Y} \}$ where $N$ is the number of the labeled data, an utility model $f:\mathcal{N}^{l} \mapsto \mathcal{Y}$, a held-out labeled testing set $\mathcal{N}^{t}=\{(x'_i,y'_i)|i\in [M], x'_i \in \mathcal{X}, y'_i \in \mathcal{Y} \}$ where $M$ is the number of the testing data, and a value function $V: (f, \mathcal{N}^{l}, \mathcal{N}^{t}) \mapsto \mathbb{R}$ (e.g., the accuracy of $\mathcal{N}^{t}$ evaluated by $f$ that is trained on $\mathcal{N}^{l}$ ). The simplest assessment metric is by performing leave-one-out (LOO) on the training set and calculating the performance differences on the testing set. The $i$-th data samples value is defined as:
\begin{align}
\label{eq:loo}
    \phi_i^{\rm LOO} = V_f(\mathcal{N}^{l}) - V_f(\mathcal{N}^{l}\backslash \{i\}).
\end{align}

A more advanced but computational costly approach is Data Shapley~\cite{pmlr-v97-ghorbani19c}. 
Shapley value for data valuation resembles a game where training data points are the players and the payoff is defined by the goodness of fit achieved by a model on the testing data. 
Given a subset $S$, let $f_{S}(\cdot)$ be a model trained on  $\mathcal{S}$. Then Shapley value of a data point $(x
_i,y_i) \in \mathcal{N}$ is defined as:
\begin{align}
\label{eq:shap}
    \phi_i^{\rm SHAP} =\mathop \sum \limits_{{S \subseteq \mathcal{N}\backslash \left\{ {x_{i} } \right\}}} \frac{{V_f\left( {S \cup \left\{ {x_{i} } \right\}} \right) - V_f\left( S \right)}}{{\left( {\begin{array}{*{20}c} {\left| \mathcal{N}^l \right| - 1} \\ {\left| S \right|} \\ \end{array} } \right)}},
\end{align}
where $V_f(S)$ is the performance of the utility model $f$ trained on subset $S$ of the data. Suppose each training of $f$ takes time $T$, the computational complexity of Eq~\eqref{eq:loo} and Eq~\eqref{eq:shap} is $\mathcal{O}(TN)$ and $\mathcal{O}(T2^N)$\footnote{In practice, there are approximation methods for calculating Shapley value, but the it still requires around $\mathcal{O}(T\rm{poly}(N))$~\cite{jia2019empirical}.}, respectively. Also, training a deep utility function (\eg neural networks) leads to a large $T$. 
 \vspace{-3mm}
\subsection{Formulation of  Unsupervised-learning-based Data Assessment }
\paragraph{Motivation story} Labeling is costly and time consuming in many medical imaging tasks. AI developers may want to pay for labeling some data points to train a particular machine learning model. In such a scenario, supervised-learning-based methods (Sec~\ref{sec:shap}) cannot fulfill the aim. Therefore, algorithms that can automatically identify low quality data before labeling data are highly desired. 

To address computational issue and demand for task/label-agnostic data quality, we propose to conduct quantitative data quality assessment via unsupervised learning. Different from the formulations of LOO~(Eq.~\eqref{eq:loo}) and Shapley value~(Eq.~\eqref{eq:shap}), here we propose a new problem formulation to infer $i$-th data's quality by assigning it a value $\phi_i : (\mathcal{X}, i) \mapsto \mathbb{R}$ using unlabeled data only.

\section{Our Method}\label{method}

\subsection{Theoretical Implication }
\label{sec:theory}



Our proposed \ours{} is well motivated by representation theory of SSL. We start by restating Theorem ~\ref{th:ssl} proved in~\cite{lee2021predicting} that under proper assumptions, the embedded space obtained by the reconstruction-based SSL strategy forms a linearly separable space of the embedded feature and a related task. Then, \textit{Remark}~\ref{th_remark} presents how we use Theorem~\ref{th:ssl} to guide the design of \ours{}.

\begin{theorem}[informal~\cite{lee2021predicting}]
\label{th:ssl}
For two views of a data $X_1, X_2 \in \mathcal{X}$ and their classification label $Y \in \bR^k$. 
Under the class conditional independence assumption, \ie $X_1 \perp X_2 |Y$, for some $w\in \bR^{m\times k}$
the representation $\psi^*: \mathcal{X} \mapsto \bR^m $ that minimizes a reconstruction loss 
${\cL(\psi)}=\E_{(X_1,X_2)}\left[\|X_1 - \psi(X_2) \|^2\right]$ 
satisfies
\begin{align*}
    w^\top \psi^*(X_1) = \E[Y|X_1].  
\end{align*}
\end{theorem}


\begin{remark}
\label{th_remark}
Theorem \ref{th:ssl} indicates two desired properties with `good' data that (approximately) satisfy class-conditional independence. First, the data will have a good geometric property in the learned representation space, namely they become clusters that are (almost) linearly separable (by $w$). Second, the learned representation has \textit{variance (top singular value of its covariance matrix)} controlled by that of $\E[Y|X_1]$, which is very \textbf{small} (since the label is almost determined entirely by the image itself). Such properties on the reconstruction-based SSL embedding space are not satisfied for `bad' data. Therefore when observing data $X_1$ is noisy or with low quality, $\psi^*(X_1)$ tends to have higher variance.  
\end{remark}
\vspace{-5mm}

\begin{figure}[t]
    \centering
    \includegraphics[width=0.95\linewidth]{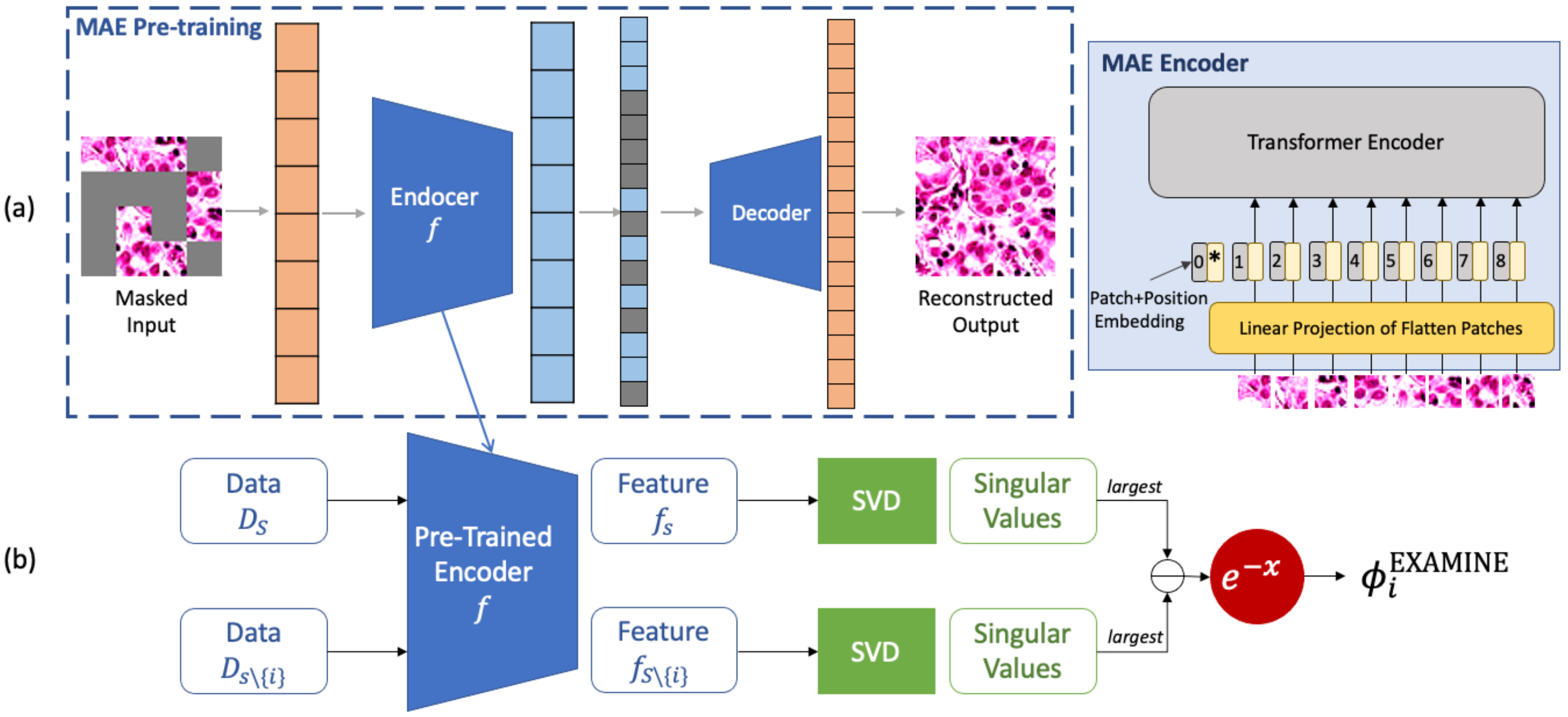}
    \caption{Proposed pipeline for \ours{} data assessment. (a) Using the state-of-the art reconstruction-based SSL strategy, MAE~\cite{he2021masked} architecture for pre-training an representation extractor (encoder). (b) \ours{} first utilizes the pre-trained encoder to extract semantic features $f_S$ and $f_{S\backslash\{i\}}$ from input data $D_S$ and $D_{S\backslash\{i\}}$, where $D_{S\backslash\{i\}}$ denotes input data $D_S$ \textit{without} data point $i$. The features then pass the SVD module to find the largest singular values $\lambda_{S}$ and $\lambda_{S\backslash\{i\}}$. The \ours{} score of data point $i$ is defined as Eq.~\eqref{eq:lorre}.}
    \label{fig:pipeline}
\end{figure}

\subsection{Data Assessment on Singular Value}



As shown in the Fig.\ref{fig:pipeline}(b), to assess data from a dataset $D_S\in R^{N\times C}$, where $N$ is the number of data and $C$ is the dimension of data, we denote the dataset without $i$-th data point as $D_{S\backslash\{i\}} \in R^{(N-1)\times C}$. To begin with, we employ SSL and the unlabeled data to train an encoder that is able to extract the low-dimensional semantic information. We denote the representation of the SSL embedding space of $D_S$ and $D_{S\backslash\{i\}}$ as $f_S$ and $f_{S\backslash\{i\}}$, respectively. Lastly, we perform SVD on both feature representations, and use the largest singular values ($\lambda_S$ and $\lambda_{S\backslash\{i\}}$) as the assessment indicator, that is, removing a `good' data point $i$ results in small change in the top singular value of embedded data representation $f$ (explained in Sec.~\ref{sec:theory}). Thus, the \ours{} score is defined as 
\begin{align}
\label{eq:lorre}
     \phi_i^{\rm EXAMINE} = \exp\left(-(\lambda_S - \lambda_{S\backslash\{i\}})\right),
\end{align}
where $\phi_i^{\rm EXAMINE} \in (0,1)$ and a larger $\phi_i^{\rm EXAMINE}$ indicates better data quality\footnote{$\lambda_S > \lambda_{S\backslash\{i\}}$ is for sure given the properties of singular value.}. Note that \ours{} is also a leave-one-out strategy but evalated on the change of the largest singular value.

We claim two advantages of using SVD for data assessment. First, by performing the SVD-based evaluation, we do not need any knowledge about the corresponding labels. This is not only the primary difference from previous methods, but also a perfect fit to our problem set up - finding good data to be labeled. Second, unlike previous methods (\ie LOO and Data Shapely, see Sec.~\ref{sec:shap}) that rely on extensively training a new model for different data combinations, 
our proposed method is efficient by performing SVD once for each data point without additional model training after the SSL encoder has been trained offline. 

\subsection{Forming Embedding Space using Masked Auto-encoding} \label{method:mae}
As the raw medical images are high-dimensional and have spurious features (\eg density, light, dose) that are irrelevant to their labels, directly applying SVD to them cannot capture task-related variance. Based on our theory developed on reconstruction-based SSL (Sec. \ref{sec:theory}), we utilize a state-of-the-art reconstructed-based strategy, Masked Auto-Encoder (MAE)~\cite{he2021masked} to learn lower-dimensional semantic feature embedding. 
As shown in Fig.~\ref{fig:pipeline}(a), MAE utilizes state-of-the-art image classification framework, Vision Transformer (ViT)~\cite{dosovitskiy2020image}, as the encoder for semantic feature extraction, and uses a lighter version of ViT as decoder. 
It first divides an input image into patches, randomly blocks a certain percentage of image patches, and then feeds them into the autoencoder architecture. 
By blocking out a large amount of image patches, the model is forced to learn a more complete representation. 
With the aim of positional embedding and transformer architecture, MAE is able to generalize the relationship between each image patch and obtain the semantic information among the whole image, which achieves the state-of-the-art performance in self-supervised image representation training. \textit{This also reduces the correlation between spurious features and labels}, compared to the traditional dimension reduction methods~\cite{chen2020self}.


\section{Experiment} \label{exp}
\subsection{Experiment setup and dataset} \label{exp:setup}
We  evaluate \ours{} on a binary classification task for PCam~\cite{veeling2018rotation},  a microscopic dataset (image size $96 \times 96$) for identifying metastatic tissue in histopathologic scans of lymph node sections. Since noise is usually the main corruption in medical images, we add non-zero mean Gaussian noise to a portion of the data to simulate real world scenario.

We split the dataset into four disjoint sets following the scale of \cite{pmlr-v97-ghorbani19c}:\\
\noindent  - \textbf{SSL Pre-Training Set} and \textbf{Assessed Set}: 160,000 and 500 unlabeled data points randomly sampled from PCam, respectively. We add 4 different level ($\mathcal{N}(\delta, \delta \times m)$, where $m$ is the mean of the dataset and $\delta=\{0.1, 0.3, 0.5, 1\}$) of noise to 60000 data points of \textbf{SSL Pre-Training Set} and 400 data points in \textbf{Assessed Set}.
    
\noindent - \textbf{Clean Train Set} and \textbf{Validation Set}: 100 and 20,000 labeled data points randomly sampled from PCam. They are used to validate the data selection in an example downstream task after obtaining \ours{} scores ($\phi_i^{\rm EXAMINE}$).

The experiments are run on NVIDIA GeForce RTX 3090 Graphics card with PyTorch. For MAE training, we select Cosine Annealing LR scheduler~\cite{loshchilov2016sgdr} and AdamW LR optimizer~\cite{loshchilov2017decoupled} with weight decay 0.05 and momentum $\{0.9, 0.95\}$. 
We train MAE for 200 epochs using batch size 256
,and set image size 72, patch size 8, 
masking ratio $40\%$. As indicated in \cite{chen2020simple, grill2020bootstrap, he2021masked} that SSL training requires a large amount of data, we begin with training a MAE on \textbf{SSL Pre-Training Set}. To ensure the training stability, we first train MAE without any noisy data, and finetune it afterward. This step is to distinguish our setting from detection out of distribution samples. After pretraining the MAE encoder, we use the frozen encoder layers as our backbone to extract the low dimensional representations. 
All of our experiments are repeated five times with different random seeds and we report the mean value of the five trials. 

\begin{figure}[t]
    \centering
    \subfloat[\small {\ours{} with MAE} ]{
        \includegraphics[width=0.28\linewidth]{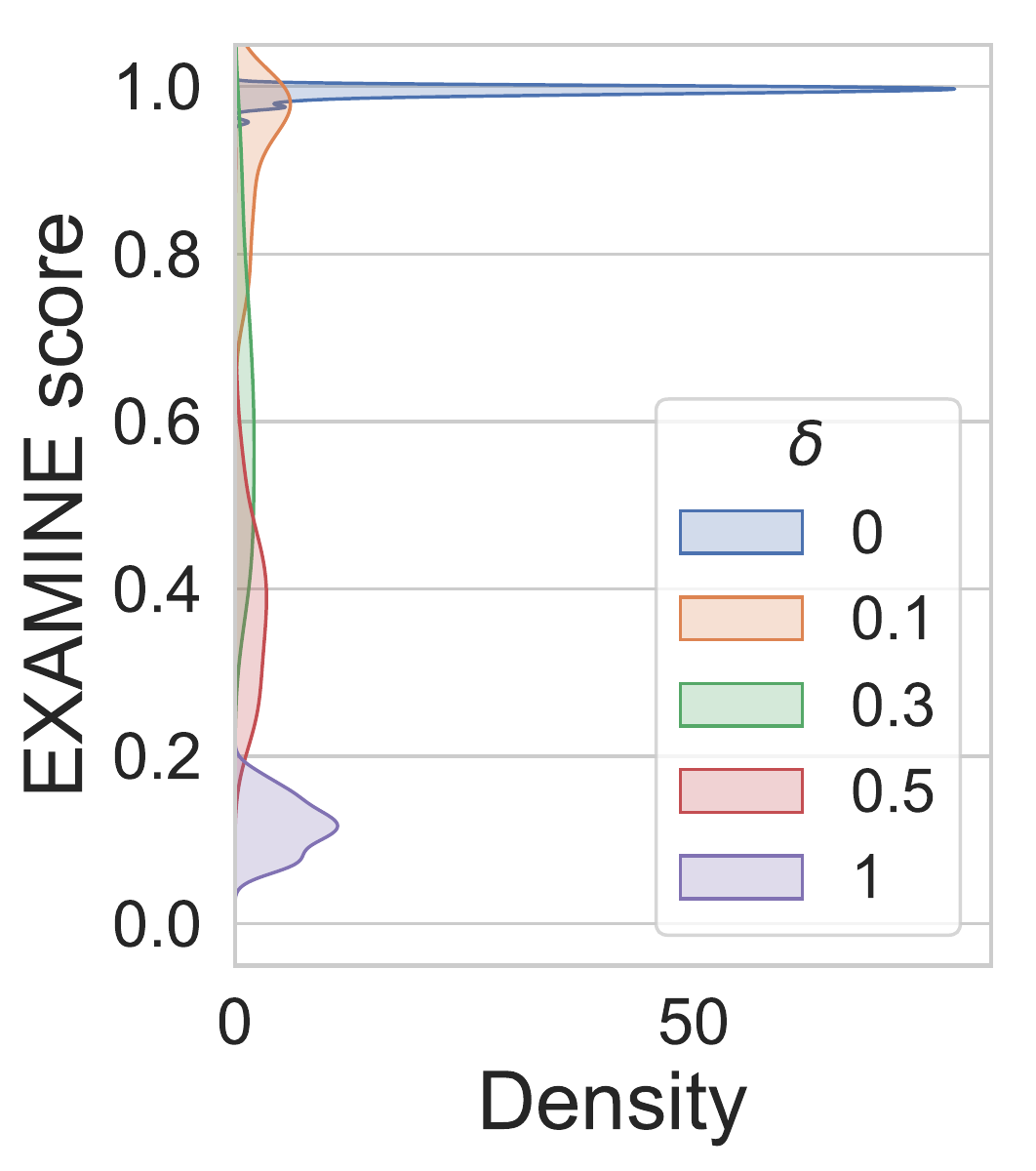}
        \label{fig:mae_noisy}
    }\qquad
    \subfloat[Comparison with different embedding]{
        \includegraphics[width=0.60\linewidth]{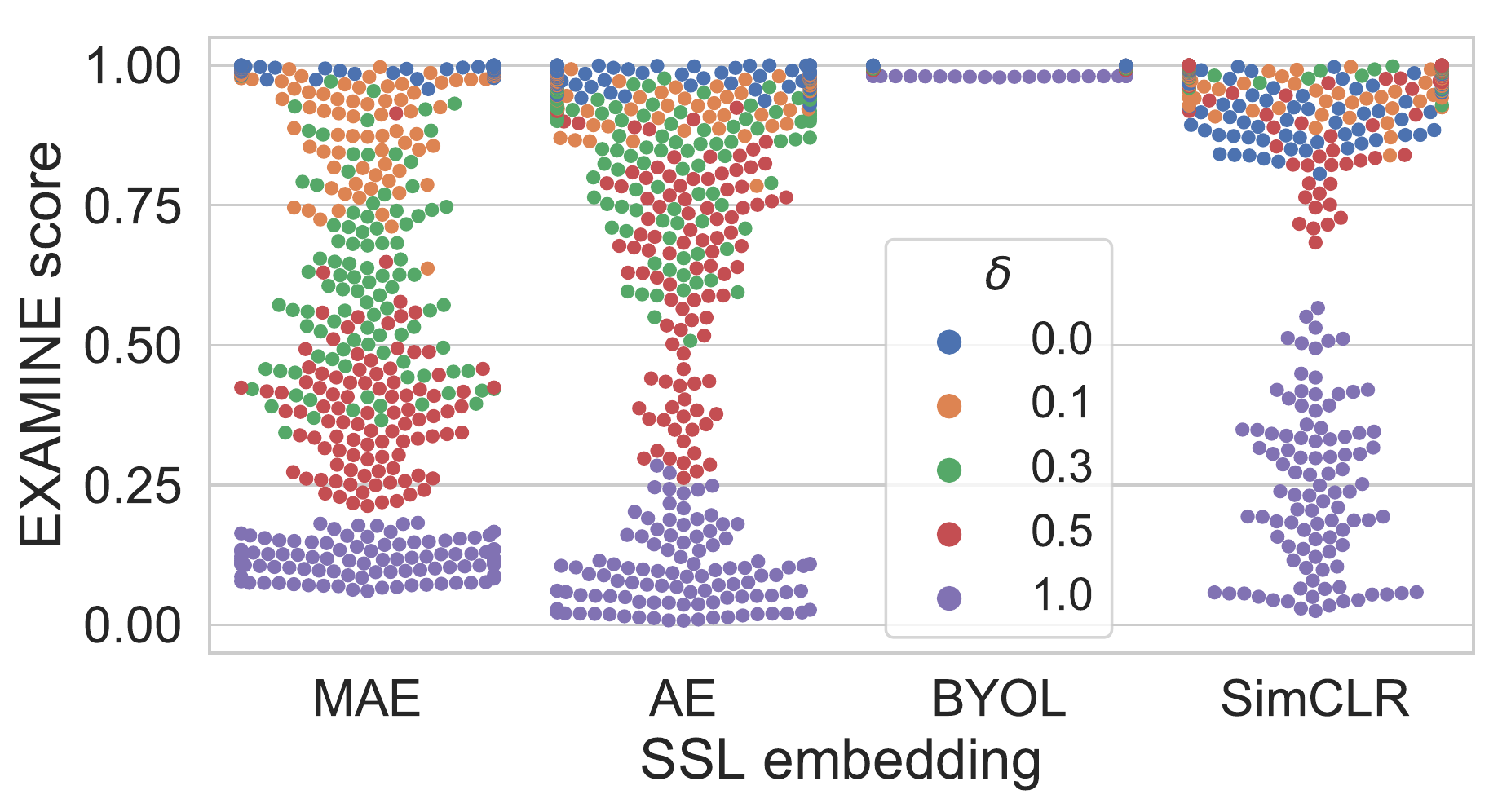}
        \label{fig:value_embed}
    }
    
    \caption{Proof of concept and comparison with baseline embedding methods. 
    Observe that \ours{} scores ($\phi_i^{\rm EXAMINE}$) get lower when noise level increases (a),
    and the \ours{} scores ($\phi_i^{\rm EXAMINE}$) in (b) shows that reconstrucion-based SSL methods perform better in separating different noise levels. }
    \label{fig:vs_ssl}
\end{figure}

\subsection{Proof of Concept with `Ground-truth'}
To validate the correctness of ranking the samples in \textbf{Assessed Set}, we plot the distributions of \ours{} scores ($\phi_i^{\rm EXAMINE}$) at different data corruption levels of \textit{noisy} setting in Fig~~\ref{fig:mae_noisy} Note that $\phi_i^{\rm EXAMINE} \in (0,1)$ and data point $i$ with larger $\phi_i^{\rm EXAMINE}$ indicates that it is considered a good data. Specifically, \ours{} score ($\phi_i^{\rm EXAMINE}$) approaching 1 indicates the difference between the top singular values are small, thus it will be considered good data. The \ours{} scores ($\phi_i^{\rm EXAMINE}$) for the high-quality data ($\delta =0$) are close to 1 and significantly higher than the corrupted data. The \ours{} scores ($\phi_i^{\rm EXAMINE}$) of data with low-level corruption ($\delta=0.5$) are also separable from those with high-level corruption($\delta=1$). 


\subsection{Comparison with Alternative Embedding Methods} \label{exp:embeddings}

We investigate the alternative feature encoders and compare their performance with MAE. Specifically, we replace MAE with SimCLR~\cite{chen2020simple} and BYOL~\cite{grill2020bootstrap}, two alternative SSL algorithms, and Autoencoder (AE)~\cite{hinton2006reducing}, a na\"{\i}ve reconstruction-based embedding strategy. 
SimCLR learns embedding by enforcing the closeness of an image and its augmented views while enlarging the distance from other images in the dataset (or batch). BYOL regularizes the multi-views of an image without sampling negative samples by training two similar networks (the online network and the target network) simultaneously. 
We use the same strategy as training MAE for these alternative encoders. 
Fig.~\ref{fig:value_embed} shows that using MAE embedding to calculate \ours{} scores ($\phi_i^{\rm EXAMINE}$) provides the best separability for the clean data from the corrupted data. The reason is that MAE best satisfies the theoretical conditions that support our proposal (Theorem~\ref{th:ssl}). AE is second to MAE, but separation boundaries are less clear.   

\begin{figure}[t]
    \centering
    \subfloat[Add high value data]{
        \includegraphics[width=0.49\linewidth]{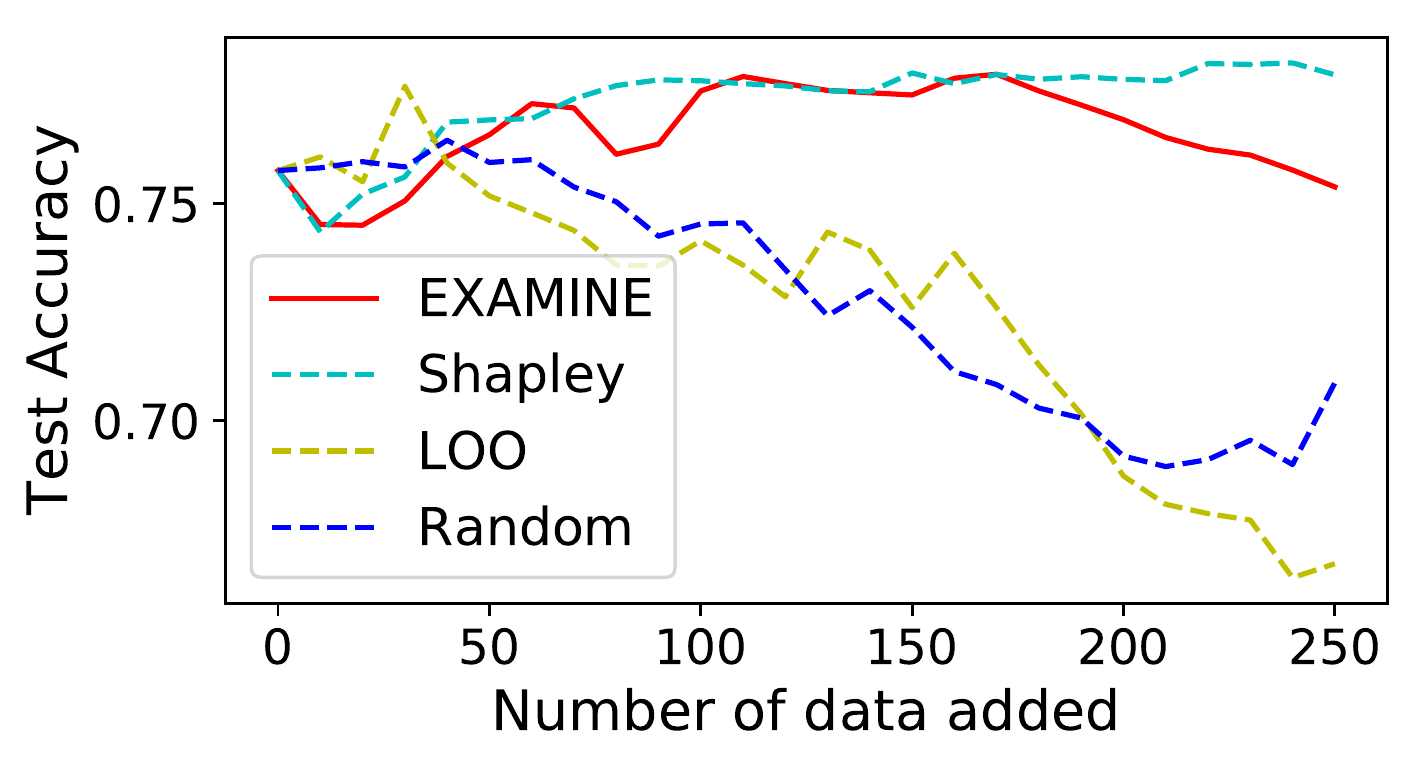}
        \label{fig:vs_valuation_add_high}
    }
    \subfloat[Add low value data]{
        \includegraphics[width=0.49\linewidth]{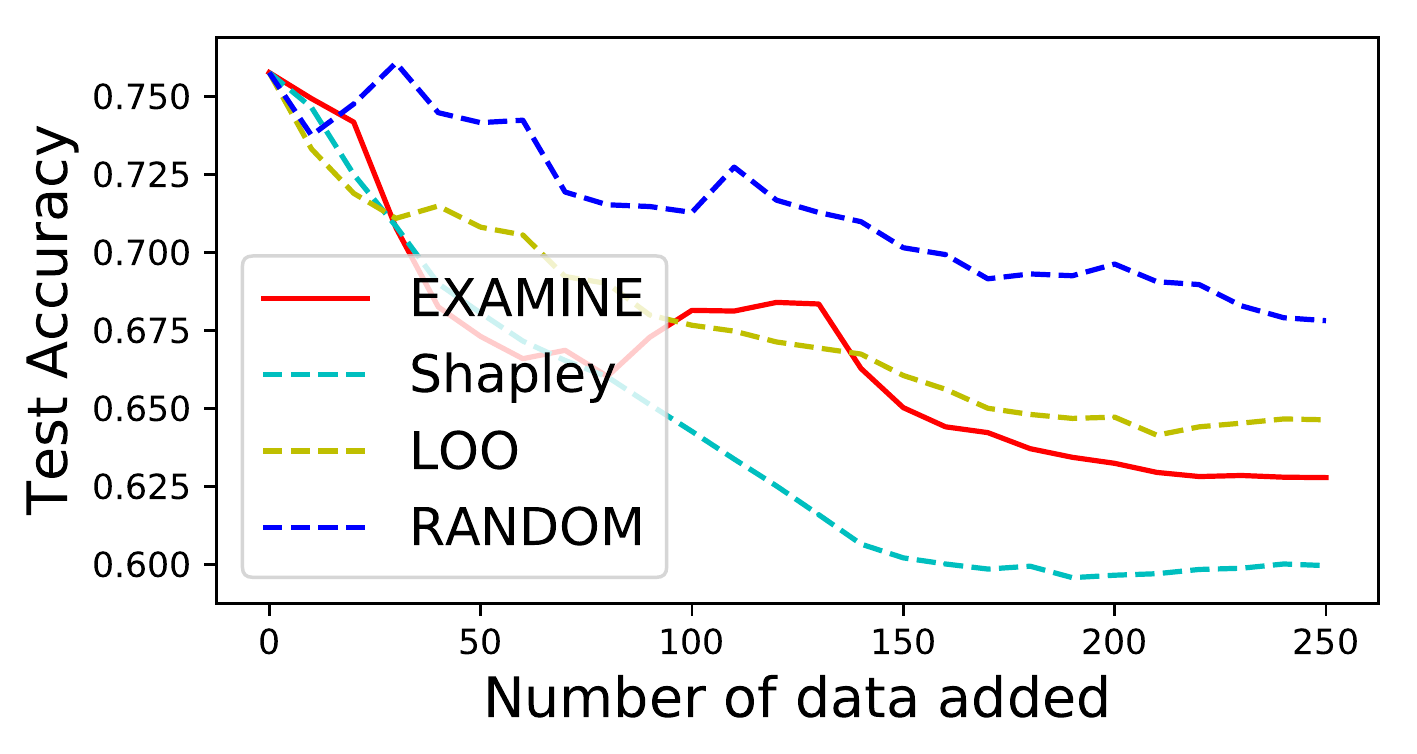}
        \label{fig:vs_valuation_add_low}
    }\\
    
    \subfloat[Remove high value data]{
        \includegraphics[width=0.49\linewidth]{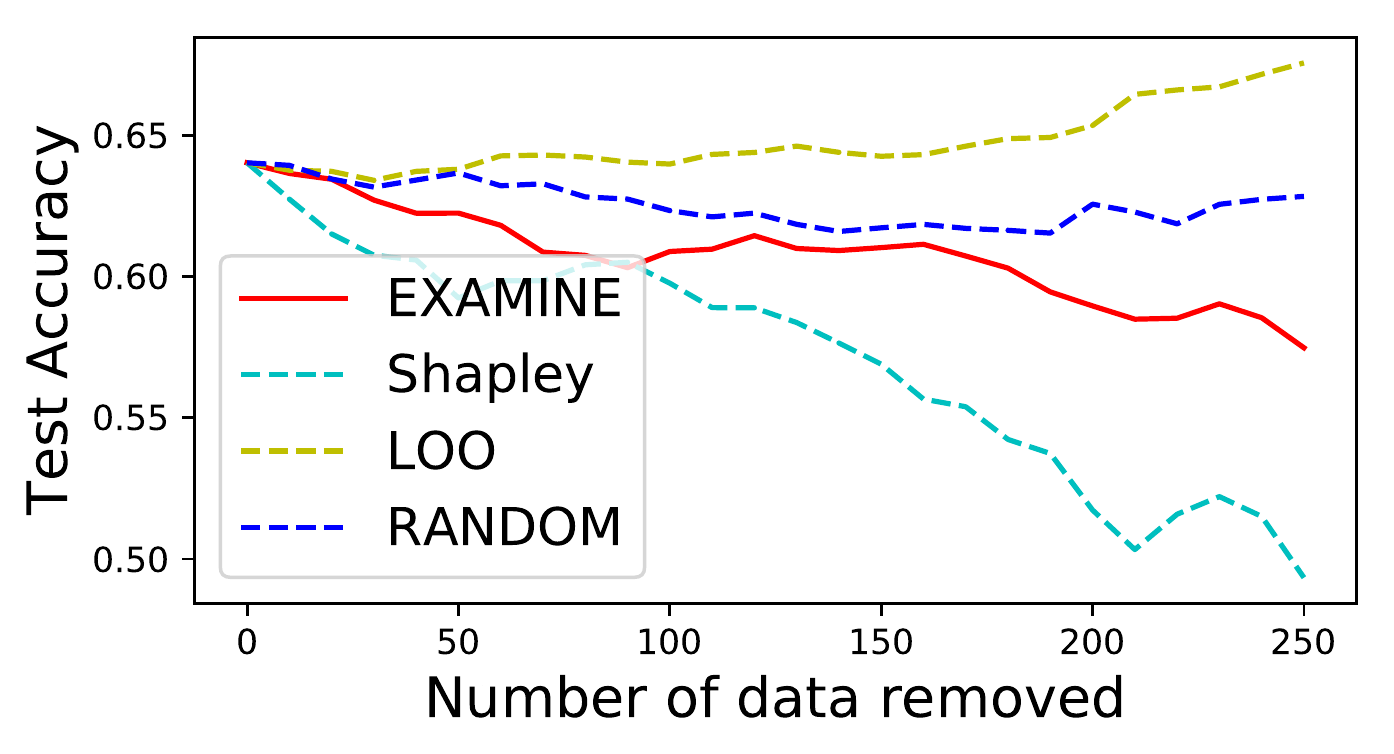}
        \label{fig:vs_valuation_remove_high}
    }
    \subfloat[Remove low value data]{
        \includegraphics[width=0.49\linewidth]{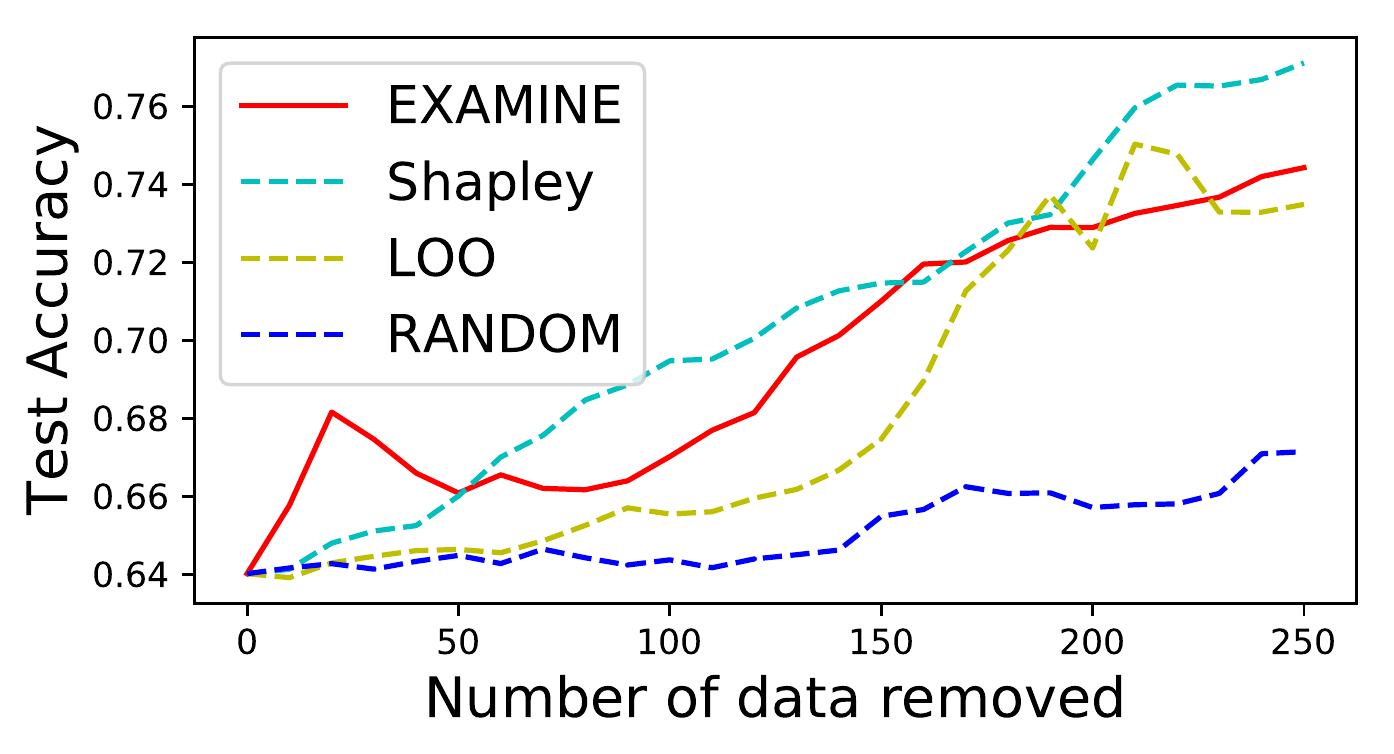}
        \label{fig:vs_valuation_remove_low}
    }
    
    \caption{Comparison with baseline data valuation methods. Adding good data(a) and removing bad data(d) should increase accuracy. Adding bad data(b) and removing good data(c) should result in accuracy drop. 
    We conclude that \ours{} can help model training by identifying good and bad data.}
    \label{fig:vs_valuation}
\end{figure}

\subsection{Comparison with Baseline Data Valuation Methods} \label{exp:data_valuation}

We compare \ours{} with supervised data valuation methods, LOO (Eq.~\ref{eq:loo} and Truncated Monte Carlo (TMC) version of Data Shapley~(Eq.~\ref{eq:shap})~\cite{pmlr-v97-ghorbani19c}, as well as a baseline method that randomly assigns data values. 
All these methods are applied on \textbf{Assessed Set}'s features extracted by the pre-trained MAE. 
We design four experiments to evaluate how selecting data using the different data assessment methods
can affect the classification accuracy. We report the averaged test accuracy on \textbf{Validation Set} using logistic regression models (LRM).

Fig.~\ref{fig:vs_valuation_add_high} and Fig.~\ref{fig:vs_valuation_add_low} show the results of adding data for training. We start with a LRM trained on the small \textbf{Clean Training Set}, and then add good/bad data from \textbf{Assessed Set} following the descending/ascending orders of their data values.
Our results show that adding data with high \ours{} score ($\phi_i^{\rm EXAMINE}$) achieves comparable accuracy curve as Data Shapley, while adding data with low \ours{} scores ($\phi_i^{\rm EXAMINE}$) results in similar curve as Data Shapley in the beginning and overall lies in between Data Shapley and LOO. This indicates that \ours{} score ($\phi_i^{\rm EXAMINE}$) is able to identify what data to be labeled and added to training set.
Fig.~\ref{fig:vs_valuation_remove_high} and Fig.~\ref{fig:vs_valuation_remove_low} show the results of removing data for training. We first train LRM on \textbf{Assessed Set}, and then remove good/bad data following the descending/ascending orders of their data values. 
Our result shows that removing high/low \ours{} score ($\phi_i^{\rm EXAMINE}$) data results in accuracy curve that is slightly worse than Data Shapley. We would like to emphasize Data Shapley uses utility function which requires labels to determine the data value, while \ours{} score ($\phi_i^{\rm EXAMINE}$) is calculated only on data itself, which is more efficient in real-world scenario. Overall, \ours{} shows comparable(or at best slightly worser) data assessment performance to Data Shapley \textit{without} knowing the labels of the data.

In addition to successfully providing correct inspection on data quality, 
our method significantly reduces the computational cost without requiring training utility functions.
The running time to obtain the data values using \ours{}, LOO, and TMC Data Shapley for the whole \textbf{Assessed Set} under our experiment setting are 23 seconds, 10 minutes, and 460 minutes, respectively\footnote{The running time for LOO and Data Shapley can significantly increase if we use a deep neural network as the utility model.}.

\section{Discussion and Conclusion}
We present a new and efficient unsupervised data evaluation method, \ours{} scores $\phi_i^{\rm EXAMINE}$, to assess data quality. With the help of MAE encoder, we can map data to the provable low-dimensional embedding space. The marginal differences on the largest singular value of data representation matrices can effectively separate data at different quality levels and achieve comparable performance with supervised data valuation methods when considering a specific task. This work takes a novel approach to promote AI in healthcare by identifying low quality data. We plan to test on larger scale medical datasets and collect domain experts' evaluations
in the future.

\section*{Acknowledgement}
This work is supported in part by the Natural Sciences and Engineering Research Council of Canada (NSERC) and NVIDIA Hardware Award. 

\bibliographystyle{splncs04}
\bibliography{ref}


\end{document}